\documentclass{article}

\usepackage{microtype}
\usepackage{graphicx}
\usepackage{subfigure}
\usepackage{booktabs} 

\usepackage{hyperref}

\usepackage[accepted]{icml2019-withoutfooter}
\usepackage{amsmath}
\usepackage{amssymb}
\usepackage{multirow}

\newcommand{\our}{DeepSubQE}
\begin{document}

\twocolumn[
\icmltitle{\our: Quality estimation for subtitle translations}




\begin{icmlauthorlist}
\icmlauthor{Prabhakar Gupta}{pv}
\icmlauthor{Anil Nelakanti}{pv}
\end{icmlauthorlist}

\icmlaffiliation{pv}{Prime Video, International Expansion, Bangalore, India}

\icmlcorrespondingauthor{Prabhakar Gupta}{prabhgup@amazon.com}
\icmlcorrespondingauthor{Anil Nelakanti}{annelaka@amazon.com}


\vskip 0.3in
]



\printAffiliationsAndNotice{}

\begin{abstract}
Quality estimation (QE) for tasks involving language data is hard owing to numerous aspects of natural language like variations in paraphrasing, style, grammar, etc.
There can be multiple answers with varying levels of acceptability depending on the application at hand.
In this work, we look at estimating quality of translations for video subtitles.
We show how existing QE methods are inadequate and propose our method \our~as a system to estimate quality of translation given subtitles data for a pair of languages.
We rely on various data augmentation strategies for automated labelling and synthesis for training. 
We create a hybrid network which learns semantic and syntactic features of bilingual data and compare it with only-LSTM and only-CNN networks. 
Our proposed network outperforms them by significant margin.
\end{abstract}

\section{Introduction}
\label{sec:intro}
Digital entertainment industry is growing multifold with ease of internet access and numerous options for on-demand streaming platforms such as Amazon Prime Video, Netflix, Hulu etc. These providers increase their viewership by enabling content in local languages. 
Translation of subtitles across languages is a preferred cost effective industry practice to maximize content reach.
Subtitles are translated using bilingual (and sometimes multilingual) translators. They watch the content and use a source subtitle in one language to translate it to another language. Low translation quality and high man-power cost which grows significantly with
scarcity of target language resources are some problems with bilingual translators. 
Low translation quality can cause increased usage drop-off and hurt content viewership for audience of target language. 
Hence, translation quality estimation (QE) is one crucial step in the process. 
Currently, a second translator evaluates the quality making evaluation as expensive as generating the translation itself.

Automated QE has been studied through the lens of binary classification between acceptable and unacceptable translation or scoring (or rating) to assign a score of translation acceptability within a given range. 
However, binary classification ignores ``loosely" translated samples that often occur due to human judgment like paraphrasing, under-translation or over-translation. 
Translators often rephrase sentences using contextual information from the video that is not available in the source sentence. 
For the scoring approach, gathering large enough sample of reliable human validated data to train a supervised system is very expensive, time-consuming and does not scale to new languages.

In this work, we propose an automated QE system \our~reducing both cost and time in subtitle translation while assuring quality. 
To overcome the problem with conventional binary approaches, we introduced a third category of translation called Loose translations. 
%
%
Our system takes a pair of sentences as input; one in source language and one in target language and classifies it in one of three categories --- \textit{Good translation}, \textit{Loose translation} or \textit{Bad translation}. This paper makes the following contributions:
\begin{itemize}
    \item We develop a novel system that can estimate quality of translations generated by either humans or MT systems with application to video subtitling. 
    \item We demonstrate achieving good generalization for subtitling QE by augmenting data with various strategies including signals from learners that themselves fail to generalize as well for the task.
    \item We present a formulation that can handle paraphrasing and other contextually acceptable non-literal translations through appropriate synthesis of \textit{Loose Translations}.
\end{itemize}

The paper is divided in following sections. Section \ref{sec:prev-work} describes the different existing methods used for QE. Section \ref{sec:approach} discusses our approach to the problem, followed by section \ref{section:datasetgeneration} which explains the details of the dataset which we generated for training. Section \ref{sec:model-arch-train} describes the model architecture and training environment. Section \ref{sec:experiments} presents the experiments, results and observations. Section \ref{sec:conclusion} concludes our findings and presents future endeavours. 

\section{Related work}
\label{sec:prev-work}
QE is important for evaluation of machine translation (MT) systems. 
Automatic evaluation of machine translation systems is a well studied topic.
Metrics like BLeU \cite{Papineni2001BleuAM} and METEOR \cite{Banerjee2005METEORAA} are some industry-wide accepted metrics to evaluate a translation where a reference text is available. 
Specifically, for each candidate translation (that is machine generated for MT systems) a reference generated by human translator is necessary for computing the metric.
However, we are interested in the setting where no reference is available.

Alternatively, HTER \cite{Snover2006ASO} is a metric used to estimate the post-edits required to improve an MT generated candidate translation to the level of human translation.
Numerous models have been proposed to predict HTER for a given MT output \cite{DBLP:conf/coling/BlatzFFGGKSU04, Specia2009EstimatingTS}.
State of the art for predicting HTER is given by a two-level neural Predictor-Estimator (PE) model \cite{DBLP:conf/wmt/KimLN17}.
PE finds the closest matching target token for each token in source and uses such matches to find the aggregate quality of translation for sentence pairs.

\begin{figure}[h]
    \includegraphics[width=\columnwidth]{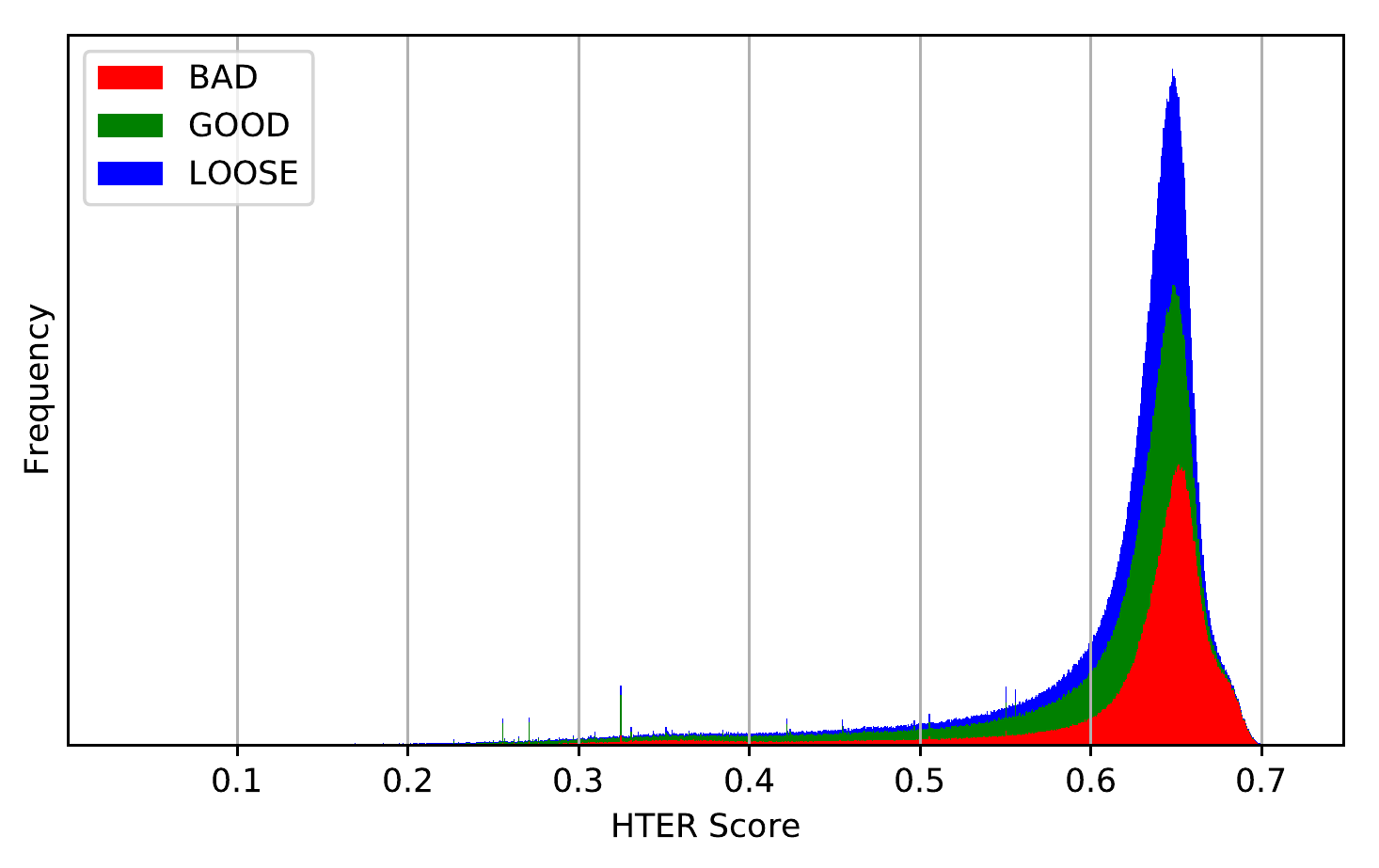}
    \caption{HTER score distribution of PE Model on our dataset computed using OpenKiwi toolkit \cite{DBLP:conf/acl/KeplerTTVM19}, see in color.}
    \label{fig:wmt-output-on-dataset}
\end{figure}

However, HTER metric is trained on data generated from MT output and hence is biased to error patterns generated by such systems.
But errors from subtitle files have a very different distribution and its patterns are not aligned with those that HTER captures. 
For example, subtitles sometimes have complete mistranslations due to alignment errors that lie outside the space of MT outputs that HTER models are trained on.
Figure \ref{fig:wmt-output-on-dataset} shows lack of separation of HTER predictions from PE model on our three class data of Good, Loose and Bad.
Hence, HTER is unsuitable for our task of evaluating subtitle quality.

Our requirement is a method that evaluates quality subtitle data that could either be human or MT generated.
The set of translations span complete mistranslations (due to issues like alignment errors), loose translations (from additional contextual information and paraphrasing) and good translations (literal translations with complete overlap of meaning).
Apart from mistranslations, errors could also arise from drift in translations, captioning of non-spoken content, etc (see~\cite{DBLP:journals/corr/abs-1909-05362} for a survey).
None of the existing methods directly apply to our problem and we work on tailoring one accordingly to our use case.
We define a three way classification to account for the three classes of translation output.
Using signals from multiple diverse methods we gather data the represents our notion of classes.
A neural network is trained on this data that classifies a given pair of subtitle blocks to one of the three classes.

\section{Approach}
\label{sec:approach}

We define the QE problem given a subtitle file $A$ in source language and its translation in target language $B$.
Quality is measured on individual pairs of translation text blocks $(A_i$, $B_i)$ that are matched by timestamps.
Each text block could contain more than one sentence spoken by multiple speakers and captions for non-spoken content (such as whispers, laughs, loudly, etc).
The problem is to assign the translation $\hat{y} \gets f(A_i, B_i)$ to one of three categories $\hat{y} \in \{\text{Good, Loose, Bad}\}$. 
A translation is \textit{Good}  if it is a perfect or near-perfect translation retaining all meaning from source and reads fluently.
It is \textit{Loose} if it is paraphrased or contains some contextual information not available in source text.
Translations of colloquial phrases and idioms also lie in this category.
\textit{Bad} translations are those in which the sentence pair have no overlap of meaning and the target is disconnected from the context in the video.


\begin{table*}[htbp]
  \centering
  \caption{Data distribution from different sources (in \%)}
    \begin{tabular}{r|c|c|c|c|c|c}
      & \textbf{Statistical} & \multirow{2}[0]{*}{\textbf{NMT}} & \textbf{Added} & \textbf{Scrambled} & \textbf{Drifted } & \textbf{Randomly} \\
      & \textbf{Classification} &   & \textbf{Captions} & \textbf{Text} & \textbf{Aligned} & \textbf{ Aligned} \\ \hline
        French & 18.83 & 33.76 & 6.58 & 6.58 & 17.13 & 17.13 \\
        German & 17.26 & 32.14 & 7.07 & 7.07 & 18.23 & 18.23 \\
        Italian & 16.44 & 32.95 & 6.57 & 6.57 & 18.74 & 18.74 \\
        Portuguese & 16.47 & 33.09 & 6.59 & 6.59 & 18.63 & 18.63 \\
        Spanish & 17.82 & 29.15 & 7.01 & 7.01 & 19.50 & 19.50 \\
    \end{tabular}%
  \label{tab:datasource-distribution}%
\end{table*}

Gathering sufficient human labelled data to train a supervised system is both expensive and time-consuming. 
Lack of suitable publicly available subtitle data for this task motivated us to reuse large volumes of unlabelled subtitles. 
We use signals like timestamp alignment and overlap statistics between source and target along with MT output for short sentences with common phrases for synthesizing samples from the three classes to learn the QE classifier.
Our augmentation methods also use statistical classifiers of lower capacity trained on external data.
The diversity and quality of data generated is critical for the QE classifier to learn good classification boundaries and generalize sufficiently well to unseen data.
We show from experiments on subtitles and other parallel corpora that by fitting a QE classifier of sufficiently high capacity to data generated as described here we get good generalization for our task.
Further details of our data augmentation methods are described in Section~\ref{section:datasetgeneration}.

\begin{figure*}[htbp]
    \includegraphics[width=\linewidth]{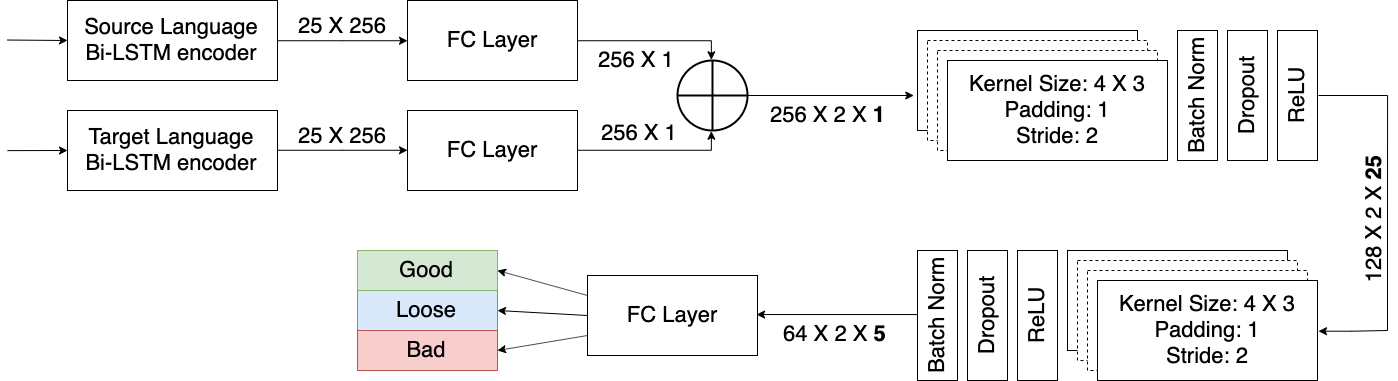}
    \caption{Visualization of \our\ model's architecture.}
    \label{fig:modelarch}
\end{figure*}

We experiment with multiple neural network architectures for the classifier beginning with simpler ones with only CNNs and LSTMs.
A hybrid architecture of Bidirectional LSTM (BiLSTM) \cite{Graves2005FramewisePC} followed by CNN outperformed them. 
Input to each model is a pair of sentences in two languages that outputs the probability of it belonging to three three classes. 
We limit the length of each sentence to 25 tokens, lowercase each word and include punctuation and numbers.  
Section~\ref{fig:modelarch} describes the model in detail.




\section{Data augmentation}
\label{section:datasetgeneration}

We begin with 30k video subtitle files in English with timestamp aligned translations into five languages (French, German, Italian, Portuguese, Spanish).
These translation text blocks are unlabeled and there is no information on which of three three QE classes they belong to.
We use various features measuring statistics of word and semantic overlap  between the source and the target as signals to label them.
These features are used with two different methods to assign two distinct scores to each sample indicating the likelihood of it being a Good or a Bad sample.
Samples where both scores are in strong agreement are given the corresponding class label of Good or Bad.
Others where both scores are aligned but only one of them is a strong indicator with high magnitude are marked as Loose.
The rest of the sample with no agreement among the scores are discarded.

\textbf{Bag-of-words model} (BOW). The first is a two parameter model that uses aligned pretrained embeddings~\cite{DBLP:journals/corr/abs-1710-04087} to score the sentence pair.
The embeddings are used create a cosine similarity matrix $S$ of size $(N \times M)$ for a source and target sentences with $N$ and $M$ words respectively.
A score $s_{\textrm{src}}$ ($s_{\textrm{tgt}}$) is computed for the source (target) by thresholding at $\theta_1$ and taking a max over the columns (rows) and averaging over the rows (columns). Correspondingly,
\begin{align}
    s_{\textrm{src}} &= \frac{1}{N} \sum_{i=1}^{N} \max_{j\in[M]}  \tau_{\theta_1}( S_{ij} ) ~\textrm{and}\\
    s_{\textrm{tgt}} &= \frac{1}{M} \sum_{i=1}^{M} \max_{j\in[N]}  \tau_{\theta_1}( S_{ij} ),
\end{align}
where $\tau_{\theta_1}(~\cdot ~)$ is element-wise thresholding at $\theta_1$.
The intuition is to aggregate similarity scores from the most relevant words of target for each word in the source and vice-versa.
The model assigns a score $s_{\textrm{BOW}} = \textrm{minimum}(s_{\textrm{src}}, s_{\textrm{tgt}})$.
The parameters are chosen by tuning on validation data of positives from NMT (implemented using ~\cite{DBLP:journals/corr/abs-1712-05690}) output and negatives from misaligned subtitles with no learning.
All samples labeled through this method are from subtitles which helps the final QE model to learn common patterns in video subtitles.

\textbf{Random Forest Classifier} (RFC).
This model uses features similar to BOW model with MUSE embeddings to train a random forest classifier on EuroParl dataset~\cite{Tiedemann2012ParallelDT}. Translations in EuroParl are augmented with errors such as incorrect word substitution and random sentence alignment to generate incorrect translations.
The model benefits from Europarl's paraphrasings that help it learn beyond literal translations.
We used about $600$k samples for each language to train and tuned parameters by cross-validation on a validation set.
RFC model assigns a probability score $s_{\textrm{RFC}}$ to each input sentence pair.

The two scores, $s_{\textrm{BOW}}$ and  $s_{\textrm{RFC}}$, both of which are in range $[0,1]$ are now used to label data to assign a label $\hat{y}$ following,

\begin{gather}
    \hat{y} =
    \begin{cases}
        \text{Bad} & \textrm{if}~ s_{\textrm{BOW}} \leq \delta_1 \wedge s_{\textrm{RFC}} \leq \delta_1, \\
        \text{Loose} & \textrm{if}~s_{\textrm{RFC}} \leq \delta_3 \wedge s_{\textrm{RFC}} \geq \delta_2, \\
        \text{Good} & \textrm{if}~s_{\textrm{BOW}} \geq \delta_4 \wedge s_{\textrm{RFC}} \geq \delta_4 \\
    \end{cases}
    \label{eq:weak-classifier-final-calc}
\end{gather}
where the thresholds are manually set to $(\delta_1,\delta_2,\delta_3,\delta_4)=(0.25,0.4,0.7,0.8)$.
This labels about $69\%$ of all samples into three classes.
All others samples with scores that do not fall in the specified ranges are filtered out due to disagreement among the models.
We refer to this set as \textit{Statistical Classification}. 
Performance and training details of BOW and RFC are reported in Appendix \ref{appendix:binary-classifiers}.
We further augment this data with ~\textit{NMT} on short sentences with frequent phrases like greetings with the Good label.
More samples for the Loose category are generated by adding captions to the source (like whispers, sighs, etc) referred as \textit{Added Captions} or changing word order in the target to degrade fluency called \textit{Scrambled Text} samples.
This constitutes all the positives (Good and Loose) while we synthesize negatives (Bad) in two ways; we randomly choose a target for each source for easy negatives (called \textit{Random Aligned}) and  to choose a target from a temporally close block for hard negatives (called \textit{Drifted alignment}).
Label distribution of the final data is reported in Table~\ref{tab:datalabel-distribution}.

\begin{table}[htbp]
  \centering
  \caption{Dataset label distribution (in \%)}
    \begin{tabular}{r|c|c|c}
      & \textbf{Bad} & \textbf{Good} & \textbf{Loose} \\ \hline
    French & 39.50 & 34.17 & 26.33 \\
    German & 38.65 & 33.06 & 28.29 \\
    Italian & 39.40 & 34.33 & 26.27 \\
    Portuguese & 39.54 & 34.10 & 26.36 \\
    Spanish & 42.05 & 29.92 & 28.03 \\
    \end{tabular}%
  \label{tab:datalabel-distribution}
\end{table}%

\section{Model architecture}
\label{sec:model-arch-train}
State-of-the-art monolingual information retrieval systems \cite{Huang2013LearningDS, Shen2014LearningSR} use a hybrid architecture of RNN followed by a convolution network to extract semantic and syntactic features of text respectively. 
We extended their idea to build a network with two monolingual encoders for source and target each to extract semantic features followed by a CNN for syntactic features. 
Refer to figure~\ref{fig:modelarch} for visualization of the network's architecture. 
Input to the model are 300 dimensional embeddings from pretrained FastText \cite{DBLP:journals/tacl/BojanowskiGJM17} for each token.
We used two BiLSTMs for each encoder with the outputs of both LSTM concatenated. 
They were then sequentially fed to two convolution modules. 
CNN output was passed through a fully connected layer before making a three class prediction. 
We used ReLU activation with dropout and Batch Normalization.
We used Adam optimizer \cite{Kingma2014AdamAM} in all our experiments with a batch size of 8192. We chose a learning rate of $10^{-3}$ scheduled to drop by a factor of $10$ twice whenever the rate of training loss drop was less than $10^{-3}$ before stopping training.
Table \ref{tab:model-performance} shows the size of the dataset used for training and testing. 

\section{Experiments}
\label{sec:experiments}
\begin{table*}[htbp]
  \centering
  \caption{Model accuracy on train and tests sets.}
    \begin{tabular}{r|r|r|r|r|r|r|r}
      & \multicolumn{2}{c|}{\textbf{Train}} & \multicolumn{5}{c}{\textbf{Test}} \\ \cline{2-8}
      & \textbf{\# Samples} & \textbf{Accuracy} & \textbf{\# Samples} & \textbf{Accuracy} & \textbf{Precision} & \textbf{Recall} & \textbf{F-Score} \\ \hline
        French & 4.23M & 93.91 & 0.83M & 91.49 & 91.04 & 90.42 & 90.63 \\
        German & 12.92M & 95.18 & 2.53M & 93.90 & 93.68 & 93.29 & 93.42 \\
        Italian & 3.74M & 94.41 & 0.73M & 92.12 & 91.64 & 91.00 & 91.24 \\
        Portuguese & 15.43M & 94.20 & 3.03M & 92.73 & 92.50 & 91.59 & 91.89 \\
        Spanish & 18.24M & 93.14 & 3.58M & 91.45 & 90.90 & 90.39 & 90.42 \\
    \end{tabular}%
  \label{tab:model-performance}%
\end{table*}%

\begin{figure}[h]
    \includegraphics[width=\columnwidth]{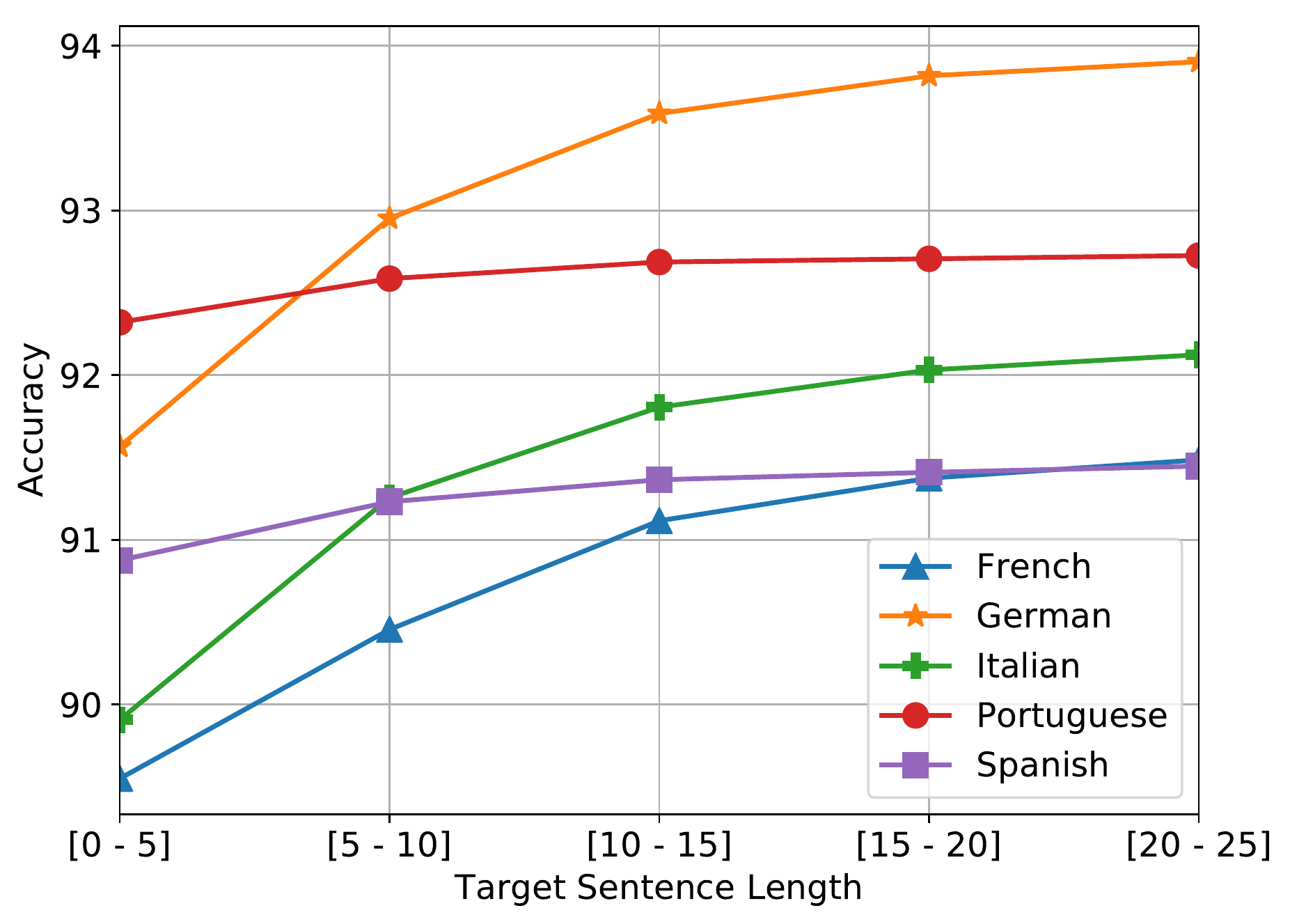}
    \caption{Model performance across target sentences of different lengths.}
    \label{fig:modelperformance-target-length}
\end{figure}

Table~\ref{tab:model-performance} shows the model comparison across various measures. We observe that the model performs similarly with accuracy of above $91\%$ for all five languages. 
The model also has similar performance across sentences of various lengths as shown in figure~\ref{fig:modelperformance-target-length} with longer sentences doing slightly better than shorter ones.
This is possibly because shorter ones when paraphrased are harder to detect than longer ones.


\begin{figure}[h]
    \includegraphics[width=\columnwidth]{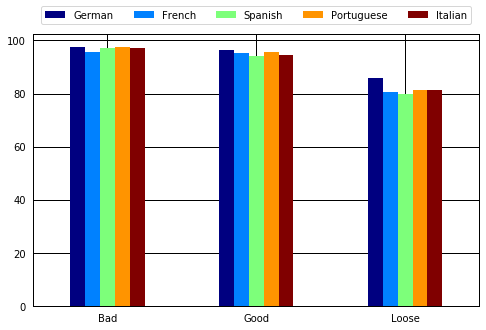}
    \caption{Model accuracy for each label, plot in color.}
    \label{fig:modelperformance-label}
\end{figure}

\subsection{Miss rate on parallel corpora}
We took a set of high quality subtitles that were translated and validated independently by two sets of distinct human translators. 
A list of parallel sentences were extracted from them using their timestamp alignment information. 
These translations that went through two rounds of human audit can safely be assumed to contain no Bad translations.
Since data consists only positives constituting of Loose and Good, we use miss rate or false negative rate (FNR) as the performance metric for this experiment. 
We ran a similar test on public EuroParl data for reference that also has only positives. 
Corresponding FNR numbers are listed in Table~\ref{tab:human-validation-results} along with number of parallel sentences in the test set. 

\begin{table}[h]
    \centering
    \caption{Miss rate on parallel corpora.}
    \resizebox{\columnwidth}{!}{%
    \begin{tabular}{r|r|r|r|r}
         & \multicolumn{2}{c|}{\textbf{High-quality subs}} & \multicolumn{2}{c}{\textbf{EuroParl}} \\ \cline{2-5}
         & \textbf{\# sentences} & \textbf{FNR} & \textbf{\# sentences} & \textbf{FNR} \\ \hline
        French & 2.4k & 13.69 & 888k & 2.69 \\
        German & 9.9k & 13.94 & 919k & 2.55 \\
        Italian & 2.4k & 10.72 & 830k & 3.40 \\
        Portuguese & 21.3k & 12.30 & 891k & 4.01 \\
        Spanish & 25.2k & 12.31 & 888k & 2.87 \\
    \end{tabular}%
    }
    \label{tab:human-validation-results}%
\end{table}

The FNR is low for all language pairs and most false negatives we identified were contextual translations that should have been marked by the model as Loose but were mistaken as Bad.
For example --- in English-German, the phrase ``\textit{Jesus}" was rewritten to ``\textit{Meine Güte.}" which literally translates to \textit{my goodness}. 
Such cases were under 14\% showing a good performance on Bad vs rest which is more critical than the Loose vs Good boundary.
Further, our subtitles data seem to have a higher sample of such contextually correct translations that are not a literal match.
Such paraphrasing is one area that we could improve upon in future work.

\subsection{Classification vs scoring}
\label{sec:class-ranking-diff}
As briefly discussed in Section~\ref{sec:intro}, QE can also be formulated as a scoring problem.
We chose the classification route using cross entropy loss for our model.
We compare an alternative that employs scoring based formulation 
extending the ordinal regression objective from \cite{Liu2018ACD} that is defined as,
\begin{gather}
    \ell = \min(0, \hat{y} - \gamma_{y})^2 + \max(0, \hat{y} - \gamma_{y}')^2 \label{eq:loss-func}
\end{gather}
where $\hat{y}$ is the predicted score, $y$ is the label, $\gamma_{y}$ and $\gamma_{y}'$ are lower bound and upper bound thresholds respectively for~$y$. The values $(\gamma_{y},\gamma_{y}')$ for each label are set to $(0,0.35)$ for class Bad, $(0.35,0.65)$ for Loose and $(0.65,1)$ for Good.


\begin{table}[htbp]
    \centering
    \caption{Accuracy comparison of classification and scoring losses.}
    \begin{tabular}{r|r|r}
         & \textbf{Classification} & \textbf{Scoring} \\ \hline
        French & \textbf{91.49} & 87.46 \\
        German & \textbf{93.90} & 90.81 \\
        Italian & \textbf{92.12} & 88.61 \\
        Portuguese & \textbf{92.73} & 88.46 \\
        Spanish & \textbf{91.45} & 87.25 \\
    \end{tabular}%
    \label{tab:class-ranking-diff}%
\end{table}%


The model assigns Good samples a score higher than those to Loose which should have been higher than those assigned to Bad.
We changed the final fully-connected layer of the model to give only one output followed by a sigmoid to bound the score in the range~$[0, 1]$.
Table \ref{tab:class-ranking-diff} shows the test accuracy of both losses. We can see that the classification loss outperformed the scoring loss by approximately $4\%$ for each language. 



\begin{table*}[htbp]
    \centering
    \caption{Test Accuracy for various model architectures.}
    \begin{tabular}{r|c|c|c|c|c|c}
     & \multirow{2}[0]{*}{\textbf{Baseline}} & \multirow{2}[0]{*}{\textbf{LSTM}} & \textbf{LASER} & \multirow{2}[0]{*}{\textbf{CNN}} & \textbf{LASER} & \multirow{2}[0]{*}{\textbf{\our}} \\
      & & & \textbf{FC} & & \textbf{CNN} & \\ \hline
    French     &  68.20 &   67.25 &     62.90 &  88.75 &      88.72 &    \textbf{91.49} \\
    German     &  70.46 &   68.63 &     61.66 &  90.88 &      90.29 &    \textbf{93.90} \\
    Italian    &  66.96 &   66.80 &     60.70 &  89.63 &      89.36 &    \textbf{92.12} \\
    Portuguese &  68.40 &   70.89 &     62.19 &  89.93 &      88.24 &    \textbf{92.73} \\
    Spanish    &  70.73 &   68.86 &     61.81 &  88.50 &      87.22 &    \textbf{91.45} \\
    \end{tabular}
    \label{tab:diff-arch-performance}
\end{table*}

    \subsection{Comparison of architectures}
\label{sec:arch-compare}
We found that the existing convolution networks \cite{Liu2018MultilingualST} that try to classify bilingual dataset are not able to learn the linguistic nuances of text and fail in many cases. However, recurrent networks that are good with parsing variable length inputs with temporal dependencies complement them. We compared our hybrid network with an LSTM network, a CNN model and models using LASER sentence embeddings \cite{DBLP:journals/tacl/ArtetxeS19}. 
For LSTM network, we concatenated the output of both BiLSTMs and fed to a fully connected layer. For CNN model, we had three convolution modules followed by a fully connected layer. We use the 1024-dimension language agnostic sentence embeddings from LASER. 
Table~\ref{tab:diff-arch-performance} compares accuracy of various models on test data.
For baseline, we used the equation \ref{eq:weak-classifier-final-calc} to generate labels for test data. 
An only-LSTM network works just about as well as the baseline but only-CNN network brings significant gains.
The convolutional layer is possibly evaluating semantic retention better ~\cite{DBLP:conf/wmt/KimLN17}.
LASER FC is a classifier trained with LASER embeddings fed into a fully connected layer that performs worse than baseline.
LASER CNN is a CNN on top of LASER embeddings and performs about as well as only-CNN.
Proposed hybrid model, however, outperformed all other networks including the CNN by $3\%$. 
The LSTM when used in combination with the CNN is consistently improving prediction accuracies across languages.
One notable observation from models trained using LASER embeddings was that LASER models took about 10 epochs on average to meet our stopping criterion while \our~model took around 34 epochs on average.



\begin{figure}[h]
    \includegraphics[width=\columnwidth]{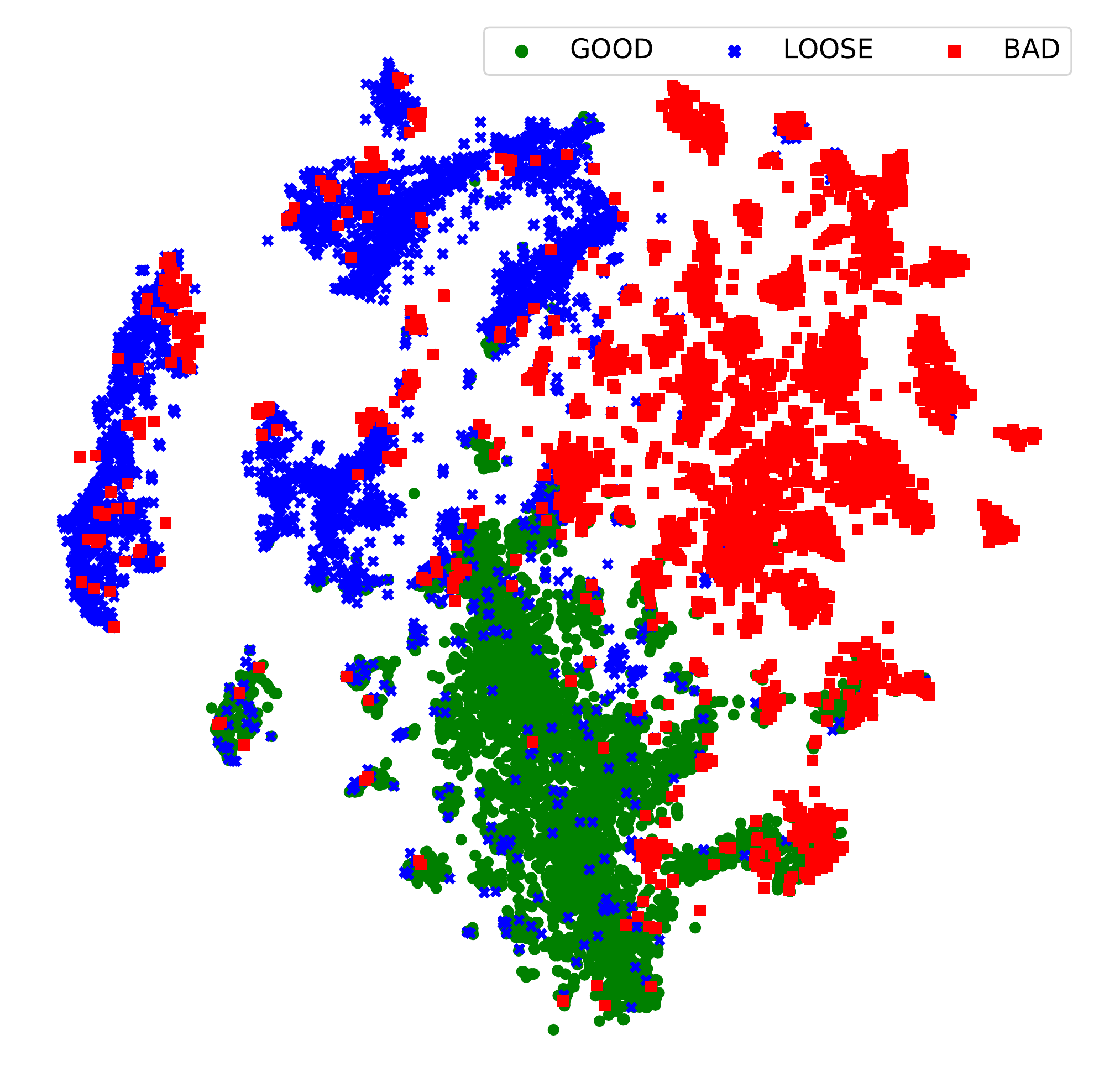}
    \caption{Visualization from t-SNE of last layer of \our.}
    \label{fig:translation-embedding-plot}
\end{figure}

In figure \ref{fig:translation-embedding-plot}, we present the t-SNE visualization \cite{Maaten2008VisualizingDU} of the last layer of our hybrid network for English-German pair.
The plot shows that there is one cluster for Good translations and one for Bad while there are two for Loose that are spread out spatially. This is because Loose translations can be close to either Good and Bad.




\section{Conclusion}
\label{sec:conclusion}


We studied the problem of translation quality estimation in video subtitles without any reference texts. 
We show, empirically, how training data can be synthesized for a three-way classification into Good, Loose and Bad translations. 
The model decision can then be integrated into the subtitle quality improvement process with Good being acceptable translations, Loose possibly requiring human post-edits and Bad needing complete rewrite. 

The current work only uses subtitle block level translations to make a decision and ignores temporal aspect. 
Temporal structure can bring significant information to make a better judgment particularly on Loose translations. 
Also, training one model per language pair requires considerable operational load. 
A multilingual model can reduce this load while helping resource starved languages. 
Exploiting temporal information and learning a common space for multiple languages are future directions we are considering for this work.


\bibliography{example_paper}

\begin{thebibliography}{22}
\providecommand{\natexlab}[1]{#1}
\providecommand{\url}[1]{\texttt{#1}}
\expandafter\ifx\csname urlstyle\endcsname\relax
  \providecommand{\doi}[1]{doi: #1}\else
  \providecommand{\doi}{doi: \begingroup \urlstyle{rm}\Url}\fi

\bibitem[Artetxe \& Schwenk(2019)Artetxe and
  Schwenk]{DBLP:journals/tacl/ArtetxeS19}
Artetxe, M. and Schwenk, H.
\newblock Massively multilingual sentence embeddings for zero-shot
  cross-lingual transfer and beyond.
\newblock \emph{Trans. Assoc. Comput. Linguistics}, 7:\penalty0 597--610, 2019.
\newblock URL \url{https://transacl.org/ojs/index.php/tacl/article/view/1742}.

\bibitem[Banerjee \& Lavie(2005)Banerjee and Lavie]{Banerjee2005METEORAA}
Banerjee, S. and Lavie, A.
\newblock Meteor: An automatic metric for mt evaluation with improved
  correlation with human judgments.
\newblock In \emph{IEEvaluation@ACL}, 2005.

\bibitem[Blatz et~al.(2004)Blatz, Fitzgerald, Foster, Gandrabur, Goutte,
  Kulesza, Sanch{\'{\i}}s, and Ueffing]{DBLP:conf/coling/BlatzFFGGKSU04}
Blatz, J., Fitzgerald, E., Foster, G.~F., Gandrabur, S., Goutte, C., Kulesza,
  A., Sanch{\'{\i}}s, A., and Ueffing, N.
\newblock Confidence estimation for machine translation.
\newblock In \emph{{COLING} 2004, 20th International Conference on
  Computational Linguistics, Proceedings of the Conference, 23-27 August 2004,
  Geneva, Switzerland}, 2004.
\newblock URL \url{https://www.aclweb.org/anthology/C04-1046/}.

\bibitem[Bojanowski et~al.(2017)Bojanowski, Grave, Joulin, and
  Mikolov]{DBLP:journals/tacl/BojanowskiGJM17}
Bojanowski, P., Grave, E., Joulin, A., and Mikolov, T.
\newblock Enriching word vectors with subword information.
\newblock \emph{{TACL}}, 5:\penalty0 135--146, 2017.
\newblock URL \url{https://transacl.org/ojs/index.php/tacl/article/view/999}.

\bibitem[Brown et~al.(1993)Brown, Pietra, Pietra, and
  Mercer]{DBLP:journals/coling/BrownPPM94}
Brown, P.~F., Pietra, S.~D., Pietra, V. J.~D., and Mercer, R.~L.
\newblock The mathematics of statistical machine translation: Parameter
  estimation.
\newblock \emph{Computational Linguistics}, 19\penalty0 (2):\penalty0 263--311,
  1993.

\bibitem[Conneau et~al.(2017)Conneau, Lample, Ranzato, Denoyer, and
  J{\'{e}}gou]{DBLP:journals/corr/abs-1710-04087}
Conneau, A., Lample, G., Ranzato, M., Denoyer, L., and J{\'{e}}gou, H.
\newblock Word translation without parallel data.
\newblock \emph{CoRR}, abs/1710.04087, 2017.
\newblock URL \url{http://arxiv.org/abs/1710.04087}.

\bibitem[Graves \& Schmidhuber(2005)Graves and
  Schmidhuber]{Graves2005FramewisePC}
Graves, A. and Schmidhuber, J.
\newblock Framewise phoneme classification with bidirectional lstm and other
  neural network architectures.
\newblock \emph{Neural networks : the official journal of the International
  Neural Network Society}, 18 5-6:\penalty0 602--10, 2005.

\bibitem[Gupta et~al.(2019{\natexlab{a}})Gupta, Sharma, Pitale, and
  Kumar]{DBLP:journals/corr/abs-1909-05362}
Gupta, P., Sharma, M., Pitale, K., and Kumar, K.
\newblock Problems with automating translation of movie/tv show subtitles.
\newblock \emph{CoRR}, abs/1909.05362, 2019{\natexlab{a}}.
\newblock URL \url{http://arxiv.org/abs/1909.05362}.

\bibitem[Gupta et~al.(2019{\natexlab{b}})Gupta, Shekhawat, and
  Kumar]{Gupta2019UnsupervisedQE}
Gupta, P., Shekhawat, S., and Kumar, K.
\newblock Unsupervised quality estimation without reference corpus for subtitle
  machine translation using word embeddings.
\newblock \emph{2019 IEEE 13th International Conference on Semantic Computing
  (ICSC)}, pp.\  32--38, 2019{\natexlab{b}}.

\bibitem[Hieber et~al.(2017)Hieber, Domhan, Denkowski, Vilar, Sokolov, Clifton,
  and Post]{DBLP:journals/corr/abs-1712-05690}
Hieber, F., Domhan, T., Denkowski, M., Vilar, D., Sokolov, A., Clifton, A., and
  Post, M.
\newblock Sockeye: {A} toolkit for neural machine translation.
\newblock \emph{CoRR}, abs/1712.05690, 2017.
\newblock URL \url{http://arxiv.org/abs/1712.05690}.

\bibitem[Huang et~al.(2013)Huang, He, Gao, Deng, Acero, and
  Heck]{Huang2013LearningDS}
Huang, P.-S., He, X., Gao, J., Deng, L., Acero, A., and Heck, L.~P.
\newblock Learning deep structured semantic models for web search using
  clickthrough data.
\newblock In \emph{CIKM}, 2013.

\bibitem[Kepler et~al.(2019)Kepler, Tr{\'{e}}nous, Treviso, Vera, and
  Martins]{DBLP:conf/acl/KeplerTTVM19}
Kepler, F., Tr{\'{e}}nous, J., Treviso, M., Vera, M., and Martins, A. F.~T.
\newblock Openkiwi: An open source framework for quality estimation.
\newblock In Costa{-}juss{\`{a}}, M.~R. and Alfonseca, E. (eds.),
  \emph{Proceedings of the 57th Conference of the Association for Computational
  Linguistics, {ACL} 2019, Florence, Italy, July 28 - August 2, 2019, Volume 3:
  System Demonstrations}, pp.\  117--122. Association for Computational
  Linguistics, 2019.
\newblock \doi{10.18653/v1/p19-3020}.
\newblock URL \url{https://doi.org/10.18653/v1/p19-3020}.

\bibitem[Kim et~al.(2017)Kim, Lee, and Na]{DBLP:conf/wmt/KimLN17}
Kim, H., Lee, J., and Na, S.
\newblock Predictor-estimator using multilevel task learning with stack
  propagation for neural quality estimation.
\newblock In Bojar, O., Buck, C., Chatterjee, R., Federmann, C., Graham, Y.,
  Haddow, B., Huck, M., Jimeno{-}Yepes, A., Koehn, P., and Kreutzer, J. (eds.),
  \emph{Proceedings of the Second Conference on Machine Translation, {WMT}
  2017, Copenhagen, Denmark, September 7-8, 2017}, pp.\  562--568. Association
  for Computational Linguistics, 2017.
\newblock \doi{10.18653/v1/w17-4763}.
\newblock URL \url{https://doi.org/10.18653/v1/w17-4763}.

\bibitem[Kingma \& Ba(2014)Kingma and Ba]{Kingma2014AdamAM}
Kingma, D.~P. and Ba, J.
\newblock Adam: A method for stochastic optimization.
\newblock \emph{CoRR}, abs/1412.6980, 2014.

\bibitem[Liu et~al.(2018{\natexlab{a}})Liu, Cui, and
  Zhao]{Liu2018MultilingualST}
Liu, J., Cui, R., and Zhao, Y.
\newblock Multilingual short text classification via convolutional neural
  network.
\newblock In \emph{WISA}, 2018{\natexlab{a}}.

\bibitem[Liu et~al.(2018{\natexlab{b}})Liu, Kong, and Goh]{Liu2018ACD}
Liu, Y., Kong, A. W.-K., and Goh, C.~K.
\newblock A constrained deep neural network for ordinal regression.
\newblock \emph{2018 IEEE/CVF Conference on Computer Vision and Pattern
  Recognition}, pp.\  831--839, 2018{\natexlab{b}}.

\bibitem[Papineni et~al.(2001)Papineni, Roukos, Ward, and
  Zhu]{Papineni2001BleuAM}
Papineni, K., Roukos, S., Ward, T., and Zhu, W.-J.
\newblock Bleu: a method for automatic evaluation of machine translation.
\newblock In \emph{ACL}, 2001.

\bibitem[Shen et~al.(2014)Shen, He, Gao, Deng, and Mesnil]{Shen2014LearningSR}
Shen, Y., He, X., Gao, J., Deng, L., and Mesnil, G.
\newblock Learning semantic representations using convolutional neural networks
  for web search.
\newblock In \emph{WWW}, 2014.

\bibitem[Snover et~al.(2006)Snover, Dorr, Schwartz, and
  Micciulla]{Snover2006ASO}
Snover, M., Dorr, B.~J., Schwartz, R., and Micciulla, L.
\newblock A study of translation edit rate with targeted human annotation.
\newblock 2006.

\bibitem[Specia et~al.(2009)Specia, Cancedda, Dymetman, Turchi, and
  Cristianini]{Specia2009EstimatingTS}
Specia, L., Cancedda, N., Dymetman, M., Turchi, M., and Cristianini, N.
\newblock Estimating the sentence-level quality of machine translation systems.
\newblock 2009.

\bibitem[Tiedemann(2012)]{Tiedemann2012ParallelDT}
Tiedemann, J.
\newblock Parallel data, tools and interfaces in opus.
\newblock In \emph{LREC}, 2012.

\bibitem[van~der Maaten \& Hinton(2008)van~der Maaten and
  Hinton]{Maaten2008VisualizingDU}
van~der Maaten, L. and Hinton, G.~E.
\newblock Visualizing data using t-sne.
\newblock 2008.

\end{thebibliography}
\bibliographystyle{icml2019}

\appendix
\section{Binary Classifiers}
\label{appendix:binary-classifiers}
In this section, we give details of the binary classifiers used for filtering data, briefly explained in Section~\ref{section:datasetgeneration}.

\subsection{Bag-of-Words Model}
We improve upon the ideas presented in IBM alignment models \cite{DBLP:journals/coling/BrownPPM94} by using aligned word embeddings \cite{DBLP:journals/corr/abs-1710-04087}. To evaluate the performance of the model, we defined another threshold ($\theta_2$) and predicted the binary label for any given translation pair following,

\begin{gather}
    \hat{y} = s_{\textrm{BOW}} > \theta_2
    \label{eq:bow-label-assignment}
\end{gather}

In table \ref{tab:dataset-performance-bow}, we report the optimum values for $\theta_1$ and $\theta_2$, dataset size and the performance of BOW model.

\begin{table}[htbp]
    \centering
    \caption{Dataset size and Performance of BOW}
    \resizebox{\columnwidth}{!}{%
    \begin{tabular}{r|c|c|c|c}
     & \textbf{$\theta_1$} & \textbf{$\theta_2$} & \textbf{Samples} & \textbf{Accuracy} \\ \hline
        French & 0.6 & 0.30 & 200k & 90.72 \\
        German & 0.6 & 0.35 & 200k & 90.51 \\
        Italian & 0.5 & 0.40 & 200k & 88.71 \\
        Portuguese & 0.6 & 0.30 & 200k & 91.89 \\
        Spanish & 0.6 & 0.30 & 200k & 90.22 \\
    \end{tabular}%
    }
  \label{tab:dataset-performance-bow}%
\end{table}%

\subsection{Random Forest Classifier}
We trained a binary Random Forest Classifier (RFC) on EuroParl dataset available in OPUS format \cite{Tiedemann2012ParallelDT}. We assumed the translations from EuroParl to be correct and introduced following errors in source text to generate incorrect translations.
\begin{itemize}
    \item \textbf{Randomly Substitute Words}: We calculate the frequency of each word from EuroParl's English corpus. Then, we remove two words with least frequency and introduce two random words at random location in sentence. By removing the least frequent words, we can try to remove more important words of the sentence and sentence loses the meaning.
    \item \textbf{Random Selected Sentence}: For every source sentence we match it with a random target sentence from the parallel corpus.
    \item \textbf{Word Trigram Substitution}: We compute the word trigram occurrence probability from EuroParl's English corpus to generate a list of possible words for any given sequence of two words. We select a trigram in sentence, replace the last word with one of the possible words list for first two words of trigram.
\end{itemize}

\begin{table}[htbp]
    \centering
    \caption{Dataset size and Performance of RFC}
    \resizebox{\columnwidth}{!}{%

    \begin{tabular}{r|c|c|c|c}
     & \textbf{Training} & \textbf{Test} & \textbf{Train} & \textbf{Test} \\    
     & \textbf{Samples} & \textbf{Samples} & \textbf{Accuracy} & \textbf{Accuracy} \\ \hline
    French & 647.5k & 161.9k & 99.95 & 92.86 \\
    German & 717.1k & 179.3k & 99.97 & 92.16 \\
    Italian & 605.2k & 151.3k & 99.93 & 92.31 \\
    Portuguese & 686.7k & 171.7k & 99.93 & 92.88 \\
    Spanish & 703.8k & 176.0k & 99.96 & 92.90 \\
    \end{tabular}%
    }
  \label{tab:dataset-performance-rfc}%
\end{table}%

We create datasets for each language maintaining the ratio of correct and incorrect translations as $1:1.2$. The sizes of train and test datasets are added in table \ref{tab:dataset-performance-rfc}. We then created a list of features, explained below, to represent a translation in a 273 length feature vector. 
\begin{itemize}
    \item \textbf{Average Vector Similarity}: Cosine similarity of average of each word's vector for each sentence.
    \item \textbf{Similarity Features}: We create a cosine similarity matrix using aligned bilingual word embeddings \cite{DBLP:journals/corr/abs-1710-04087}. We borrow the ideas from \cite{Gupta2019UnsupervisedQE}, instead of selecting a threshold and calculating the percentage of word matches in cosine similarity matrix; we take the maximum values for each column and row. We repeat process for ngrams for $n = [1,6]$ by taking the vector average for $n>2$.
    
    \item \textbf{n-gram Frequencies}: Vector of source and target unigram, bigram and trigram probabilities for each ngram in source and target sentence. 
    \item \textbf{Structural Features}: Number of words in source sentence and target sentence

\end{itemize}

\begin{figure}[htbp]
    \includegraphics[width=\columnwidth]{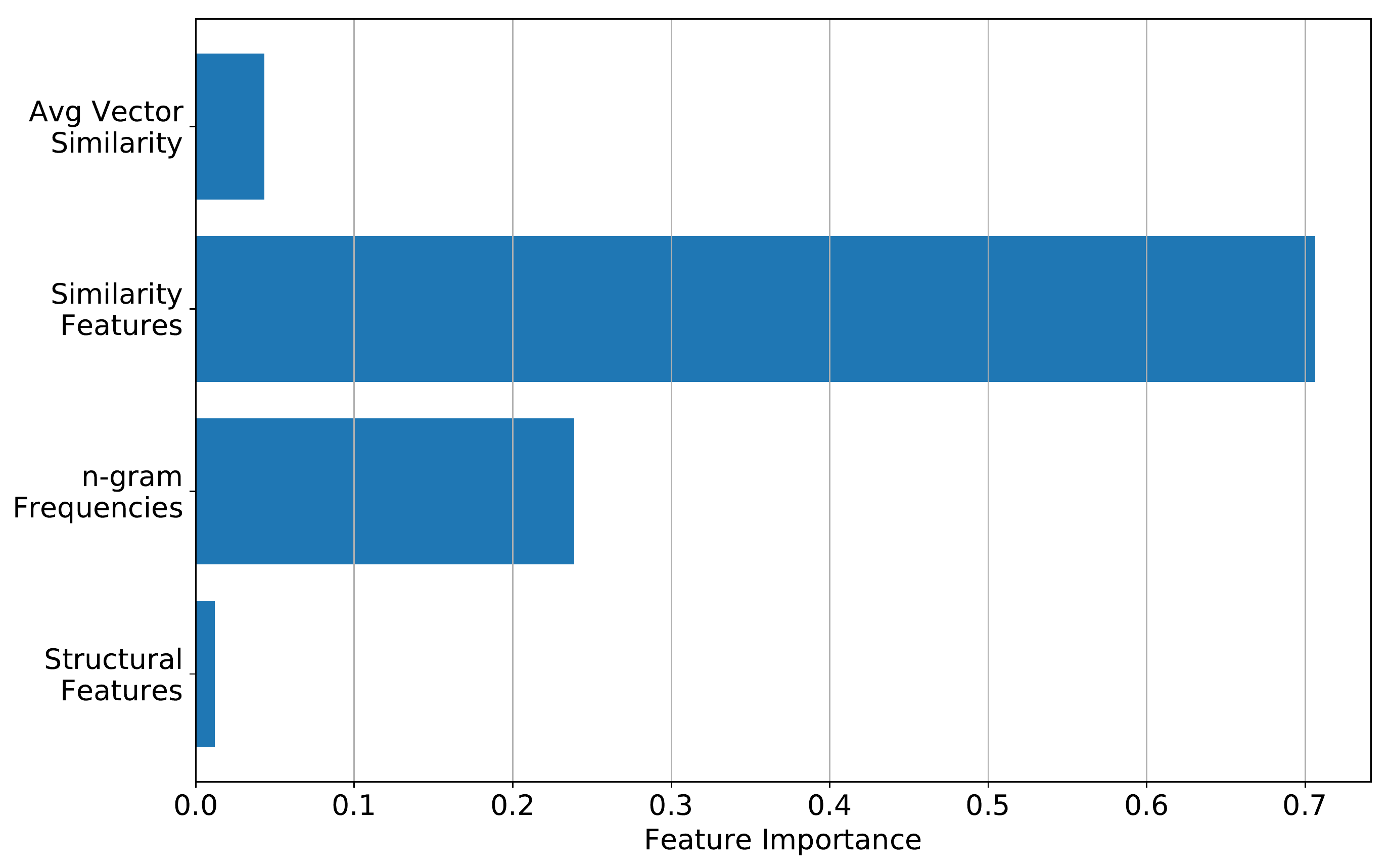}
    \caption{Feature Importance for RFC (Target-Language: German)}
    \label{fig:rfc-feature-importance-german}
\end{figure}

In table \ref{tab:dataset-performance-rfc}, we show the results for the classifier and in figure \ref{fig:rfc-feature-importance-german} show the importance of each feature type for RFC.


\end{document}